\documentclass[11pt,a4paper]{article}
\usepackage[hyperref]{emnlp2020}
\usepackage{times}
\usepackage{latexsym}
\usepackage{amsmath, amssymb}
\usepackage{breqn}
\usepackage{tabularx}
\usepackage{graphicx}
\usepackage{booktabs, makecell}
\usepackage{footnote}
\usepackage{adjustbox}
\usepackage{multirow}
\usepackage{algorithm}
\usepackage[noend]{algpseudocode}

\usepackage{url}

\usepackage{microtype}

\aclfinalcopy 
\def\aclpaperid{2280} 

\title{{A}dversarial {G}rammatical {E}rror {C}orrection}

\author{Vipul Raheja \hspace{0.5cm} Dimitris Alikaniotis\\
  Grammarly \\
  \texttt{firstname.lastname@grammarly.com} \\}

\date{5th October, 2020}

\begin{document}
\maketitle
\begin{abstract}

Recent works in Grammatical Error Correction (GEC) have leveraged the progress in Neural Machine Translation (NMT), to learn rewrites from parallel corpora of grammatically incorrect and corrected sentences, achieving state-of-the-art results.
At the same time, Generative Adversarial Networks (GANs) have been successful in generating realistic texts across many different tasks by learning to directly minimize the difference between human-generated and synthetic text.
In this work, we present an adversarial learning approach to GEC, using the generator-discriminator framework.
The generator is a Transformer model, trained to produce grammatically correct sentences given grammatically incorrect ones.
The discriminator is a sentence-pair classification model, trained to judge a given pair of grammatically incorrect-correct sentences on the quality of grammatical correction.
We pre-train both the discriminator and the generator on parallel texts and then fine-tune them further using a policy gradient method that assigns high rewards to sentences which could be true corrections of the grammatically incorrect text.
Experimental results on FCE, CoNLL-14, and BEA-19 datasets show that Adversarial-GEC can achieve competitive GEC quality compared to NMT-based baselines.
\end{abstract}

\section{Introduction}
\label{intro}
Grammatical Error Correction (GEC) has grown into a popular NLP task that deals with building systems for automatically correcting errors in written text \cite{ng-etal-2013-conll, ng-etal-2014-conll}.
Evolving from the approaches of building error-specific machine learning classifiers \cite{Tetreault:2008:UDP:1599081.1599190, DeFelice:2008:CAP:1599081.1599103, tetreault-etal-2010-using, dahlmeier-ng-2011-grammatical, TACL384}, it has gained popularity as a monolingual Machine Translation (MT) problem, where the system learns to ``translate'' a given erroneous text to its corrected form \cite{brockett-etal-2006-correcting, felice-etal-2014-grammatical, susanto-etal-2014-system}. Initially, Statistical phrase-based Machine Translation (SMT) techniques were successfully applied to the task \cite{yuan-felice-2013-constrained, junczys-dowmunt-grundkiewicz-2016-phrase, yuan-etal-2016-candidate} as as a way to handle all error types concurrently. More recently, several Neural Machine Translation (NMT) systems have been developed with promising results \cite{sutskever2014seq, bahdanau14nmt, cho-etal-2014-learning}, and their successful application to GEC, either in combination with SMT models \cite{chollampatt2016ngec, yuan-briscoe-2016-grammatical, yannakoudakis-etal-2017-neural, grundkiewicz-junczys-dowmunt-2018-near}, or strictly as neural models, has emerged as the new state-of-the-art \cite{Xie2016NeuralLC, schmaltz-etal-2017-adapting, sakaguchi-etal-2017-grammatical, ji-etal-2017-nested, ge-etal-2018-fluency, junczys-dowmunt-etal-2018-approaching, chollampatt2018mlconv, chollampatt-ng-2018-neural, DBLP:journals/corr/abs-1903-00138}.

Despite the successes of NMT-based models for GEC, a major challenge still lies in the definition of the evaluation metrics. 
Ideally, the metric should be able to quantify the (a) lexical overlap, (b) semantic similarity, and (c) grammaticality of a generated sentence, given a grammatically incorrect input sentence.
In a straightforward application of NMT-based models to the GEC task, one would minimize a surrogate loss (e.g., cross-entropy), which is an upper bound on the true loss, and hence a loose approximation of these complex criteria. 
Moreover, NMT-based GEC models try to maximize n-gram or edit-based metrics, such as $M^{2}$ \cite{dahlmeier-ng-2012-better}, $I$-Measure \cite{felice-briscoe-2015-towards}, or GLEU \cite{napoles2015ground} pushing the NMT-based models to generate sentences with n-gram precisions as high as possible, which may not necessarily lead to high-quality generation for the GEC task.
In order to avoid these issues, we take a different approach, inspired by Generative Adversarial Networks (GANs) \cite{Goodfellow:2014:GAN:2969033.2969125}, which provide a framework that can be leveraged to directly model the task based on the differences in the input-output distributions and the complex criteria mentioned above. 
Moreover, GANs have shown remarkable ability to generate coherent and semantically meaningful text in many natural language processing tasks such as machine translation \cite{pmlr-v95-wu18a,  yang-etal-2018-improving}, dialogue generation \cite{li-etal-2017-adversarial}, and abstractive summarization \cite{AAAI1816238, wang-lee-2018-learning} among others.

We propose a GAN-based generator-discriminator framework for grammatical error correction.
The generator is a Sequence-to-Sequence (Seq2Seq) model, which is trained to ``translate'' a grammatically incorrect sentence to its grammatically correct rewrite. 
The discriminator, a deep neural sentence-pair classification model is trained to evaluate the probability of the generated sentence being a lexically-similar, meaning-preserving, and grammatically correct rewrite of the incorrect input sentence. 
Adversarial training between the two models is set up as optimizing a \textit{min-max} objective, where the discriminator learns to distinguish whether a given input is sampled from the ground-truth (\textit{human}-generated) or generator (\textit{artificially}-generated) distributions, \textit{maximizing} the difference between them. The generator, on the other hand, learns to trick the discriminator by producing high-quality correction candidates, thus, \textit{minimizing} the difference between its output and a ground-truth corrected sentence. 
Further, the discriminator is used to fine-tune the generator using a policy gradient \cite{williams1992reinforce, yu2017seqgan, pmlr-v95-wu18a}, rewarding high quality generated text when conditioned on the source, improving, thus, the generation results.
By minimizing the difference between the human- and the artificially-generated distribution, we aim at directly optimizing the task based on the criteria mentioned above. 

We evaluate the effectiveness of our approach on three standard datasets on the task, observing that the discriminator can provide reasonably consistent guidance to the generator and further help improve its performance. 
Experimental results indicate that our model can achieve significantly better performance than strong NMT-based baselines.

\noindent In summary, we make the following contributions:
\begin{itemize}
\item This work is, to the best of our knowledge, the first to apply generative adversarial training to the GEC task.
\item We propose a sentence-pair classification-based discriminator, that can better distinguish grammatical text from ungrammatical text by learning to directly optimize the task rather than constructing or relying on n-gram or edit-based metrics. We analyze different formulations of the discriminator, and provide insights into how its setup, pre-training and integration into the framework can be leveraged for stable training and better performance.
\item We conduct extensive experiments on standard GEC datasets and evaluate the system against strong baselines, showing that the proposed model consistently achieves better results in a self-contained single model setting, without relying on any resources other than just the training data.
\end{itemize}

\section{Adversarial GEC}
\begin{figure}
    \centering
    \includegraphics[width=\linewidth]{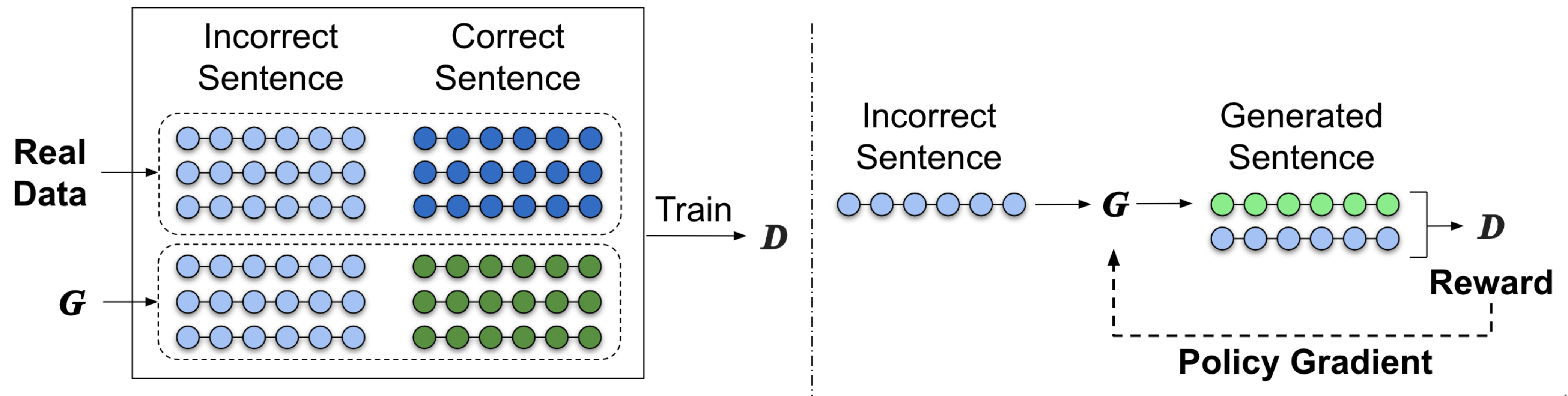}
    \caption{Adversarial-GEC training. \textbf{Left}: \(D\) is trained over the real and the generated data by a pre-trained \(G\).
    \textbf{Right}: \(G\) is further trained by policy gradient where the final reward is provided by \(D\) and is passed back to the generator.}
    \label{fig:framework}
\end{figure}

Fig. \ref{fig:framework} outlines our approach which consists of two components the (a) Generator ($G$) and (b) Discriminator ($D$).


\subsection{Generator}
\label{intro:generators}
Following recent NMT-based state-of-the-art GEC systems, we treat a grammatically incorrect sentence as the source and its grammatically corrected counterpart as the target.
Formally, given a sequence $x  = [x_1, x_2, \dots, x_S]$, we aim to generate another sequence $y = [y_1, y_2, ..., y_T]$ which is the grammatically corrected form of $x$. We denote a pair of incorrect-correct sentences as $(x, y)$. 
Given a sequence \(x\), the generator learns to produce another sequence \(y^{\prime} \approx y\).

While the generator can be any Seq2Seq model, we use two common Encoder-Decoder architectures for GEC; an attention-based RNN \cite{luong-etal-2015-effective} and a Transformer \cite{DBLP:conf/nips/VaswaniSPUJGKP17}. 

\subsection{Discriminator}
\label{sec:discriminators-sent-pair}
In this framework, a critical component is a discriminator that is responsible for providing the appropriate reward to the generator based on the quality of the generated text. Most GAN architectures typically use a single-sentence \textit{real}-vs-\textit{fake} classifier\footnote{In this context, \textit{fake} would be a (grammatically) incorrect and \textit{real} a (grammatically) correct sentence.} as the discriminator \cite{yu2017seqgan}. 
However, we argue that such a formulation does not accurately express the GEC task objective.
A conventional GAN discriminator would provide the probability of a sentence being grammatically correct as the reward.
However, it would be especially harder for such a classifier to differentiate between a ground-truth correction and a generated sentence that fits the distribution of real-world text and is far from the generated data, but does not make the intended corrections or changes the semantics of the source. Moreover, it would also be unable to provide a proportionate reward to a partially corrected sentence. Due to the lack of contextual knowledge about \textit{what} has been corrected, such a classifier would struggle to differentiate between low-quality or unsuitably corrected sequences. Consequently, it will end up giving them rewards comparable to sentences which are truly the corrected forms of given incorrect source sentences.

In the GEC task, we ultimately want the generator to generate corrected sentences that fit the constraints mentioned in Section \ref{intro}.
Hence, we formulate the objective of the discriminator as being two-fold: first, to be able to evaluate the quality of the generated text in terms of its validity compared to the ground-truth distribution, and second, to measure its quality as the appropriate rewrite for a given input sentence.
In summary, the discriminator needs to be able to measure the degree of  ``grammatical correctness'' of an output sentence, given its corresponding input sentence, instead of only distinguishing between \textit{real}-vs-\textit{fake}
Therefore, instead of training a single-sentence classifier, we train on incorrect-correct sentence pairs. We consider ground-truth data $(x, y)$ as high-quality corrections (positive examples), while data sampled from the generator $(x, y^{\prime})$ as low-quality (negative examples).
We experiment with two discriminator models for both the single-sentence and sentence-pair formulations: CNN- and RNN-based due to their simplicity, widespread use in sentence-pair modeling tasks, and ease of implementation. 

\subsection{Adversarial Training} 
Adversarial training between \(G\) and \(D\) (parameterized by $\theta$ and $\phi$, respectively) is set up as optimizing a \textit{min-max} objective, formulated as the following objective function $V(G_\theta, D_\phi)$:

\begin{dmath}
\label{adversarial-minmax}
\mathop{\min_{\theta}} \mathop{\max_{\phi}} V(G_\theta, D_\phi) \\
\begin{aligned}
= & \mathbb{E}_{(x,y) \sim P_{data}} [\log D_\phi(x, y)] + \\ 
& \mathbb{E}_{x \sim P_{data},\,y' \sim P_{G_\theta{(\cdot|x)}}} [\log(1 - D_\phi(x, y'))]
\end{aligned}
\end{dmath}

\noindent where $P_{data}$ is the underlying training data distribution and $P_{G_\theta{(\cdot|x)}}$ is the distribution of the generator output.

With this objective function, the discriminator learns to predict whether a given sentence pair has been sampled from the ground-truth data $(x,y)$ or from $G_\theta$: $(x,y^\prime)$. \(G\)$_\theta$ tries to confuse \(D\)$_\phi$ by generating high-quality corrected samples $y^{\prime} \approx y$, given a  ground-truth input sentence $x$. Formally, the objective function of \(D_\phi\) is defined as the standard binary cross entropy (BCE) loss:

\small
\begin{dmath}
\label{dis-loss}
\mathcal{L}_d = \mathbb{E}_{(x,y) \sim P_{data}} \log D_\phi(x, y) + \mathbb{E}_{x \sim P_{data}, y^\prime \sim P_{G_\theta(\cdot|x)}} \log(1 - D_\phi(x, y^\prime))
\end{dmath}
\normalsize

The objective of the generator can be formulated as optimizing the following loss:

\begin{dmath}
\small
\label{gen-loss}
\mathcal{L}_g = \mathbb{E}_{x \sim P_{data}, y^\prime \sim P_{G_\theta(\cdot|x)}} \log(1 - D_\phi(x, y^\prime))
\end{dmath}

However, since the generator performs discrete sampling to obtain $y^\prime$, we cannot directly use the gradient-based approach to backpropagate the gradients, making $V(G_\theta, D_\phi)$ non-differentiable with respect to $\theta$.
To address this issue, borrowing from \citet{cai-wang-2018-kbgan} and \citet{pmlr-v95-wu18a}, 
we use single-sample based REINFORCE \cite{williams1992reinforce}, a Monte-Carlo policy gradient method to optimize $G_\theta$. 
In Reinforcement Learning (RL) terms, the generator $G_\theta$ acts as the agent under the policy $G_\theta(\cdot|x)$, and the generated grammatically corrected sentence $y^{\prime}$ is the action.
The environment is characterized via the input sequence $x$ and the discriminator $D_\phi$, which provides the reward $- \log(1-D_\phi(x, y'))$ based on the discriminative loss of $D_\phi(x, y')$. 
The generator improves itself by maximizing the reward returned from the environment.
The gradients $\nabla_{ \phi} \mathcal{L}_d$ and $\nabla_{ \theta} \mathcal{L}_g$ can thus be estimated by sampling a correction from the generator $y^\prime \sim G(\cdot|x)$ as follows:

\small
\begin{align}
\nabla_{ \phi} \mathcal{L}_d &= \nabla_{ \phi} \log D_\phi(x, y) + \nabla_{ \phi} \log(1 - D_\phi(x, y^{\prime})) \\
\label{gen-loss-gradient-rl}
\nabla_{\theta} \mathcal{L}_g &= \nabla_{\theta} \log G_\theta(y^{\prime}|x) \log(1 - D_\phi(x, y^{\prime}))
\end{align}
\normalsize
\noindent where $\phi$ and $\theta$ can be updated as per the REINFORCE algorithm.


\subsection{Training Strategies}
\label{training-strategies}
While REINFORCE provides a framework where the reward function does not have to be differentiable, the discrete reward space due to the use of a single sampled $y^{\prime}$ to perform the Monte Carlo estimation leads to the problem of high variance, resulting in unstable training - a widely acknowledged limitation of RL methods. 
In practice, we find that adversarially training the generator solely with Eq. \ref{gen-loss} is unstable, even when it is pre-trained.
This is due to the sparsity of the rewards provided to the generator, which happens only once it has fully generated a sentence. This is also compounded by the fact that we do not generate multiple samples for computational efficiency. Hence, the generator training becomes brittle and finds it extremely difficult to get out of bad local minima or mode collapse. 
To alleviate this issue, we leverage the following measures: baseline reward, and teacher forcing/interleaved training to train the generator. 

\textbf{Baseline Reward}
A popular technique to alleviate the variance issue is the subtraction of baseline values from the original rewards. The baseline reward could be computed using various approaches. \citet{yang-etal-2018-improving} use a constant value, \citet{Rennie_2017_CVPR} use the reward of the sequence obtained by the current model with a greedy sampling strategy, \citet{DBLP:journals/corr/RanzatoCAZ15}, \citet{DBLP:conf/iclr/BahdanauBXGLPCB17}, and \citet{Liu2017ImprovedIC} use an MLP to estimate the baseline reward. However, these methods rely on approximating the terminal reward using intermediate states, or incorporating word-level rewards via rollout strategies for better credit assignment. Moreover, such approaches have been found to be extremely time-consuming, given the large decoding space. 
Based on prior works on RL for modeling dialog systems, which also have discrete action-reward spaces \cite{sankar-ravi-2019-deep, su2015reward}, we use a moving average of the historical reward values as the baseline, which stabilizes the training process and is computationally tractable.


\textbf{Interleaved Training} Following \citet{AAAI1816360} and \citet{pmlr-v95-wu18a}, we interleave MLE and Policy Gradient training. This combination of an adversarial objective with \textsc{MLE} is an important factor in successfully training $G$. 
By some probability $\lambda$ (more details in Section \ref{combining-mle-adversarial}), randomly chosen mini-batches are trained with the Policy Gradient (discriminator reward), while other mini-batches are trained using MLE. 
This alternation improves training stability, as MLE acts as a regularizer to ensure a smoother model update, alleviating the negative effects brought by high gradient estimation variance of the one-step Monte Carlo sample in REINFORCE. After this generator update, it is used to generate more realistic corrections, which are then used to train the discriminator. 
This approach is equivalent to the teacher forcing step in \citet{li-etal-2017-adversarial} and \citet{yang-etal-2018-improving}, where, after every iteration of policy gradient training update, they update the generator using teacher forcing by making the discriminator automatically assign a reward of 1 to the ground-truth data, which is used by the generator to further update itself. 

\section{Experiments}

\subsection{Data}
\label{sec:data}
In line with previous works, we use the public NUCLE corpus (used in the CoNLL 2014 GEC Shared Task \citep{ng-etal-2014-conll, dahlmeier-etal-2013-building}), the FCE Corpus \cite{DBLP:conf/acl/YannakoudakisBM11}, the Lang-8 Corpus of Learner English \cite{tajiri-etal-2012-tense}, and the Write \& Improve and LOCNESS (W\&I+L) dataset from the BEA 2019 Shared Task \cite{bryant-etal-2019-bea, Granger1998TheCL}, as our parallel training datasets.
We use CoNLL-2013 \cite{ng-etal-2013-conll}, FCE-dev, and BEA19-dev as our development sets, and for our test splits,\footnote{We could not use JFLEG \cite{napoles2015ground} corpus for evaluation due to licensing restrictions.} we use the FCE-test, CoNLL-2014 \cite{ng-etal-2014-conll} test, and the BEA19 test set (evaluated by ERRANT \cite{bryant-etal-2017-automatic}). We report $F_{0.5}$ scores evaluated by the $M^{2}$ scorer \cite{dahlmeier-ng-2012-better} for both of these test datasets.

\begin{table}[t!]
\begin{center}
\small
\begin{tabular}{llll}
\toprule
\textbf{Split} & \textbf{Dataset} & \textbf{Sentences} & \textbf{Tokens} \\
\toprule
\multirow{4}{*}{Train} & FCE-train & 27\textit{k}& 454\textit{k}\\
    & BEA19-train & 34\textit{k} & 628\textit{k}\\
    & CoNLL14-train & 57\textit{k} & 1.1M\\
    & Lang-8 & 1M & 13M\\
\midrule
\multirow{3}{*}{Dev} & CoNLL13 & 1.3\textit{k} & 28\textit{k}\\
    & FCE-dev & 1.9\textit{k} & 28\textit{k}\\
    & BEA19-dev & 4.3\textit{k} & 87\textit{k}\\
\midrule
\multirow{3}{*}{Test} & CoNLL14-test & 1.3\textit{k} & 30\textit{k}\\
    & FCE-test & 2.4\textit{k}& 36\textit{k}\\
    & BEA19-test & 4.4\textit{k} & 85\textit{k}\\
\bottomrule
\end{tabular}
\caption{Dataset splits and sizes.}
\vspace{-1.5em}
\label{tab:train_data}
\end{center}
\end{table}

\subsection{Baselines}
\label{baselines}
We use the two generators introduced in Section \ref{intro:generators} as baseline generators. 
Building on these baselines, we develop GAN frameworks, in combination with the following setups of discriminators - a) \textsc{SS}: CNN- and RNN-based Single Sentence classifier,\footnote{A failed formulation using language models is described in Section \ref{appendix-lm}.} and
b) \textsc{SP}: CNN- and RNN-based Sentence-Pair classifier (Section \ref{sec:discriminators-sent-pair}). 
We also experiment with using the GLEU score directly as the reward for an input-output sentence pair. This setting overlaps with the work of \citet{sakaguchi-etal-2017-grammatical}.\footnote{We are unable to provide comparison against \citet{sakaguchi-etal-2017-grammatical} because they report results on JFLEG Corpus. See Section \ref{sec:data} for details.}

\subsection{Implementation Details}
\subsubsection{Data}
Following \citet{junczys-dowmunt-etal-2018-approaching}, we use byte-pair encoding (BPE) sub-word units \cite{sennrich-etal-2016-neural}, which is also the way to address the issue of  out-of-vocabulary words. The vocabulary is based on 35k most frequent BPE subword units, where both the source and target side use the same vocabulary. 

\subsubsection{Generators}
We refer to \citet{junczys-dowmunt-etal-2018-approaching} for our training setup, who laid out extensive guidelines for adapting NMT-based models for the GEC task. For the RNN-based generator, following \citet{luong-etal-2015-effective}, we use 4 layers of bi-directional GRUs in both the encoder and decoder. We set the word embedding size to 512, size of hidden units for both encoder and decoder as 1024. 
For the Transformer, following the \textsc{BASE} model in \citet{DBLP:conf/nips/VaswaniSPUJGKP17}, we set up the model architecture with the encoder and decoder both having a stack of six layers of self-attention/feed-forward sub-layers. 
The word embedding size is set to 512, and the number of attention heads to 8. 
The size of the inner layer in the position-wise feed-forward network is set to 2048. 
In order to discourage copying \cite{10.5555/3157096.3157211, junczys-dowmunt-etal-2018-approaching, grundkiewicz-etal-2019-neural} we use strong dropout for regularization: layer dropout of 0.3 for both the RNN and Transformer models, attention dropout of 0.1, and source and target word dropout of 0.2 and 0.1 respectively. These hyperparameters were chosen as prescribed in the referred works, but also worked well in practice when tuned on the development sets. 

\subsubsection{Sentence-Pair Discriminators}
The RNN-based discriminator model is set up as a siamese network, sharing the same embeddings and weights, each processing one of the two sentences. The RNN-based model, for each sentence in the pair, consists of a word embedding layer of size 300, followed by two layers of bi-directional GRU, with hidden size of 128. There are residual connections at each time step between the layers. The bi-directional outputs of the last recurrent layer of both the sentences in the pair are concatenated, and used as input to a dense feed-forward layer with an output of size 128, followed by a sigmoid. We use dropout on the recurrent units and between layers (both with probability 0.2). For the CNN-based discriminator, we use the convolutional matching model used by \citet{pmlr-v95-wu18a} since \citet{10.5555/2969033.2969055} found it to have a superior performance to the siamese architecture.

\subsubsection{Training}
\label{training-details}

\begin{algorithm}[t!]
\caption{Adversarial-GEC}
\small
\begin{algorithmic}[1]
\State Initialize \textbf{$G_\theta$}, \textbf{$D_\phi$} with random weights $\theta$, $\phi$.
\State Pre-train \textbf{\textit{$G_\theta$}} on ground-truth dataset $\mathcal{D} = (X, Y)$ with MLE loss
\State Generate negative samples $\mathcal{D'} = (X, Y^\prime)$ using $G_\theta$ for training $D_\phi$
\State Pre-train $D_\phi$ on $\mathcal{D}$ and $\mathcal{D'}$ until initial accuracy $\varepsilon$ with BCE loss
\While{not converged}
\State Sample $(X, Y) \sim P_{data}$ 
\State Sample $Y^\prime \sim G_\theta{(\cdot|X)}$
\State Sample $\rho \sim [0,1]$ to determine interleaving
\If {$\rho \leq \lambda$}
    \State Compute Rewards $R$ for $(X, Y^\prime)$ using $D_\phi$
    \State Update $G_\theta$ via Policy Gradient using $R$
\Else 
    \State Update $G_\theta$ via teacher-forcing using MLE
\EndIf
\State Train $D_\phi$ using Eqn. 2, on $(X, Y)$ and $(X, Y^\prime)$
\EndWhile \\

*Parameter update equations for $G_\theta$ and $D_\phi$ are as follows: \\
$\theta \leftarrow \theta - \alpha_g \nabla_{\theta_G}$ \\
$\phi \leftarrow \phi - \alpha_d \nabla_{ \phi_D}$

\end{algorithmic}
\end{algorithm}

\begin{savenotes}
\begin{table*}
\small
\centering
\resizebox{\linewidth}{!}{%
\begin{tabular}{lccc|ccc|ccc}
\toprule
    & \multicolumn{3}{c}{FCE} & \multicolumn{3}{c}{CoNLL14} & 
    \multicolumn{3}{c}{BEA19} \\ 
    & P & R & F$_{0.5}$ & P & R & F$_{0.5}$ & P & R & F$_{0.5}$ \\ 
\midrule
\midrule 
\textbf{Baselines} \\
RNN  & 58.50 & 20.85 & 42.97 &  60.37 & 18.74 & 41.80 & 49.21 & 34.44 & 45.32\\
Transformer & 60.87 & 25.03 & 47.30 & 63.98 & 21.52 & 45.88 & 50.38 & 35.43 & 46.45 \\
\midrule
\textbf{Adversarial-GEC (Our System)} \\
RNN + CNN & \textbf{64.21} & 22.46 & 46.81 & 59.31 & 21.01 & 43.46 & \textbf{54.21} & 34.37 & 48.6 \\
Transformer + CNN & 62.53 & \textbf{27.82} & \textbf{50.04} & \textbf{64.68} & \textbf{22.57} & \textbf{47.10} & 53.78 & \textbf{36.52} & \textbf{49.13}\\
\midrule
\midrule 
\textbf{Recent GEC Systems}\\
\citet{ji-etal-2017-nested}$^{\dagger}$ & - & - & - & - & - & 41.53 & - & - & - \\
\citet{grundkiewicz-junczys-dowmunt-2018-near}$^{\dagger\dagger}$ & - & - & - & \textbf{66.61} & 17.58 & 42.76 & - & - & - \\ 
\citet{chollampatt2018mlconv}$^{\ddagger, \dagger}$ & - & - & - & 59.68 & 23.15 & 45.36 & - & - & - \\
\citet{DBLP:journals/corr/abs-1903-00138}$^{\mathparagraph}$ & - & - & - & 55.96 & 30.73 & 48.07 & - & - & - \\
\citet{kaneko2020encoderdecoder} & \textbf{61.7} & \textbf{46.4} & \textbf{57.9} &  59.2 & \textbf{31.2} & \textbf{50.2} & \textbf{51.5} & \textbf{43.2} & \textbf{49.6}\\ 
\bottomrule
\end{tabular}
}
\caption{Results of Adversarial-GEC against single-model NMT baselines of state-of-the-art GEC systems.\\
$^{\dagger}$Trained on non-public CLC data, 
$^{\dagger\dagger}$Trained on NUCLE and Lang-8, 
$^{\ddagger}$MLConv - single model,
$^{\mathparagraph}$Trained on One-Billion Word Benchmark}
\label{results:main}

\end{table*}
\end{savenotes}

A major challenge with GANs is that the joint training between the generator and the discriminator needs to be carefully coordinated, in order to stabilize the training \cite{yu2017seqgan, li-etal-2017-adversarial, yang-etal-2018-improving, pmlr-v95-wu18a, fedus2018maskgan, wang-lee-2018-learning}. Therefore, we first pre-train the generator model $G_\theta$ using maximum likelihood estimation (MLE) on the ground-truth training dataset until convergence. This stage is essential to enable the joint training to converge later, since the action space during generation is immense and applying Policy Gradient training from scratch would lead to slow and unstable training. The pre-trained model is then used to decode the training data $x$ using beam search (size 4), and generate the output sentences $y'$, essentially building the negative examples in the training data for the discriminator $(x, y')$. The discriminator is initially pre-trained on a combination of the ground-truth parallel data $(x, y)$ and the machine-generated data $(x, y')$, where $y'$ is sampled from the pre-trained generator model. The discriminator is trained until the classification accuracy reaches $\varepsilon$ (further analysis in Section \ref{discrminator-pretrain-discussion}). Once the generator and the discriminator have been pre-trained, they are adversarially co-trained, where the generator is trained with a combination of MLE and Policy Gradient (and teacher forcing), until the performance of $G_\theta$ does not improve on the development set.\footnote{More details in Appendix A.}


\section{Results}

In contrast to related works on Neural GEC, we do not use a lot of the heuristics that most recent systems leverage in order to enhance their model performance pre- and post-training. These heuristics include using spellcheckers to correct spelling errors in the data, pre-trained language models trained on large quantities of external data, synthetic data generation, re-ranking systems to sort the outputs of the generator model, among others. We chose to keep our framework simple compared to most contemporary works in that we do not leverage anything beyond what the raw training data and the baseline architectures have to offer, which makes it simple and self-contained. This decision was in the interest of system complexity, training time, and clear evaluations. The goal of this work is not to build a state-of-the-art GEC system but to demonstrate the value of adversarial training. Hence, we report results in a single-model setting, without the use of any external data or resources beyond the training data. 


The results of Adversarial-GEC compared to baseline models are presented in Table \ref{results:main}.\footnote{While multiple recent works based on pre-trained Transformers such as \citet{kiyono-etal-2019-empirical} and \citet{omelianchuk-etal-2020-gector} have pushed the state-of-the-art in GEC, they are not comparable to our work because of the use of pretrained LMs, ensembles and synthetic training data.}
These results are based on the best performing (on the development set) parameters $\varepsilon = 0.7$, $\lambda = 0.4$ using the CNN sentence-pair discriminator.
The results demonstrate a substantial improvement in $F_{0.5}$ for both adversarially trained models, across all evaluation datasets. Overall, the RNN model achieves greater gains on precision than the Transformer, which achieves greater gains on recall. We carry out statistical significance tests with bootstrap resampling, and correcting for multiple comparisons, obtain significant gains over the baselines (\textit{p} $<$ 0.01).


As mentioned in Sections \ref{sec:discriminators-sent-pair} and \ref{baselines}, we experiment with three discriminator formulations (\textsc{SS}, \textsc{SP}, \textsc{GLEU}) in the Adversarial-GEC setting to provide the rewards to guide the generators. 
Table \ref{discriminators} describes the results of using the two kinds of discriminators in each formulation (CNN, RNN) of the discriminative task, and doesn't show a significant difference in either formulation.

\section{Discussion}
In this section, we describe experimental results on adversarial training strategies, based on validation data splits. There are three parts to making the training work (a) formulating the discriminator task to compute the reward, (b) reducing the variance in rewards for better gradient estimation, and (c) combining the MLE and Adversarial objectives for more stable training. 

\subsection{Discriminator Formulation}

\begin{table}[t!]
\resizebox{\columnwidth}{!}{%
\small
\centering
\begin{tabular}{llccc}
\toprule
& Generator & FCE & CoNLL14 & BEA19 \\ 
\midrule
\midrule
\multicolumn{3}{l}{\textbf{\textsc{SS}: Single-Sentence Discriminator}} \\
\textit{CNN} & RNN & 41.68 & 40.23 & 45.53\\ 
             & Transformer & 43.45 & 41.52 & 46.31\\
\textit{RNN} & RNN & 41.21 & 39.25 & 45.58\\ 
             & Transformer & 41.36 & 39.84 & 46.86\\
\midrule
\midrule
\multicolumn{3}{l}{\textbf{\textsc{SP}: Sentence-Pair Discriminator}} \\
\textit{CNN} & RNN & 46.81 & 43.46 & 48.6 \\
             & Transformer & \textbf{50.04} & \textbf{47.10} & \textbf{49.13} \\
\textit{RNN} & RNN & 46.45 & 43.17 & 48.11 \\ 
             & Transformer & 49.88 & 46.95 & 49.02 \\
\midrule
\textit{GLEU} & RNN & 43.35 & 42.1 & 46.68\\
              & Transformer & 45.65 & 45.9 & 47.84\\
\bottomrule
\end{tabular}
}
\caption{Impact of training different Discriminator task formulations and models on $F_{0.5}$ test splits.}
\label{discriminators}
\end{table}

We observe in Table \ref{discriminators} that the single-sentence discriminator (SS) performs the worst against all discriminator formulations. Furthermore, SS performs even worse than the baseline generators, which points to the direction that it acts as a barrier in their ability to generalize.

We attribute this performance limitation to two factors. First, since the model does not consider the original sentence, it lacks the ability to learn the parts of the sentence which make it ungrammatical, rewarding similarly marginally correct and highly incorrect sentences. We investigate this idea by feeding the discriminator incorrect sentences sampled from \(P_{data}\) and observe that they get nearly the same reward from SS despite their varying degrees of incorrectness. This impedes generator improvement as any inaccuracies are penalized disproportionately. Secondly, producing grammatically correct sequences is not enough to solve the task. A generated sequence can be grammatically correct, albeit semantically or lexically different. A discriminator which lacks the contextual information provided by the original sentence can reward such sequences with a high reward propagating such false starts. Therefore, a generator that produces only \textit{one} grammatical sentence would receive a high reward from the discriminator.

On the other hand, GLEU achieves better performance compared to SS but weaker when compared to SP. This corroborates the above argument as GLEU, essentially being a special case of the SP formulation, is able to provide higher quality reward since it tries to account for fluency \textit{and} grammaticality in evaluation on references. \textsc{SP}, on the other hand, is able to go beyond the GLEU score’s low-level n-gram matching criteria, learning latent characteristics of the GEC task and providing a more appropriate reward to the generator. Acting in this way provides a much smoother objective compared with GLEU since the latter is quite sensitive to slight translation differences at the word or phrase level.
Second, the generator and discriminator co-evolve. The dynamics of the discriminator make the generator grow in an adaptive way rather than controlled by a fixed evaluation metric such as GLEU, achieving better distributional alignment, which is further verified by its superior performance.

\subsection{Balancing Discriminator Pre-Training}
\label{discrminator-pretrain-discussion}

\begin{figure}
\centering
\includegraphics[width=\linewidth]{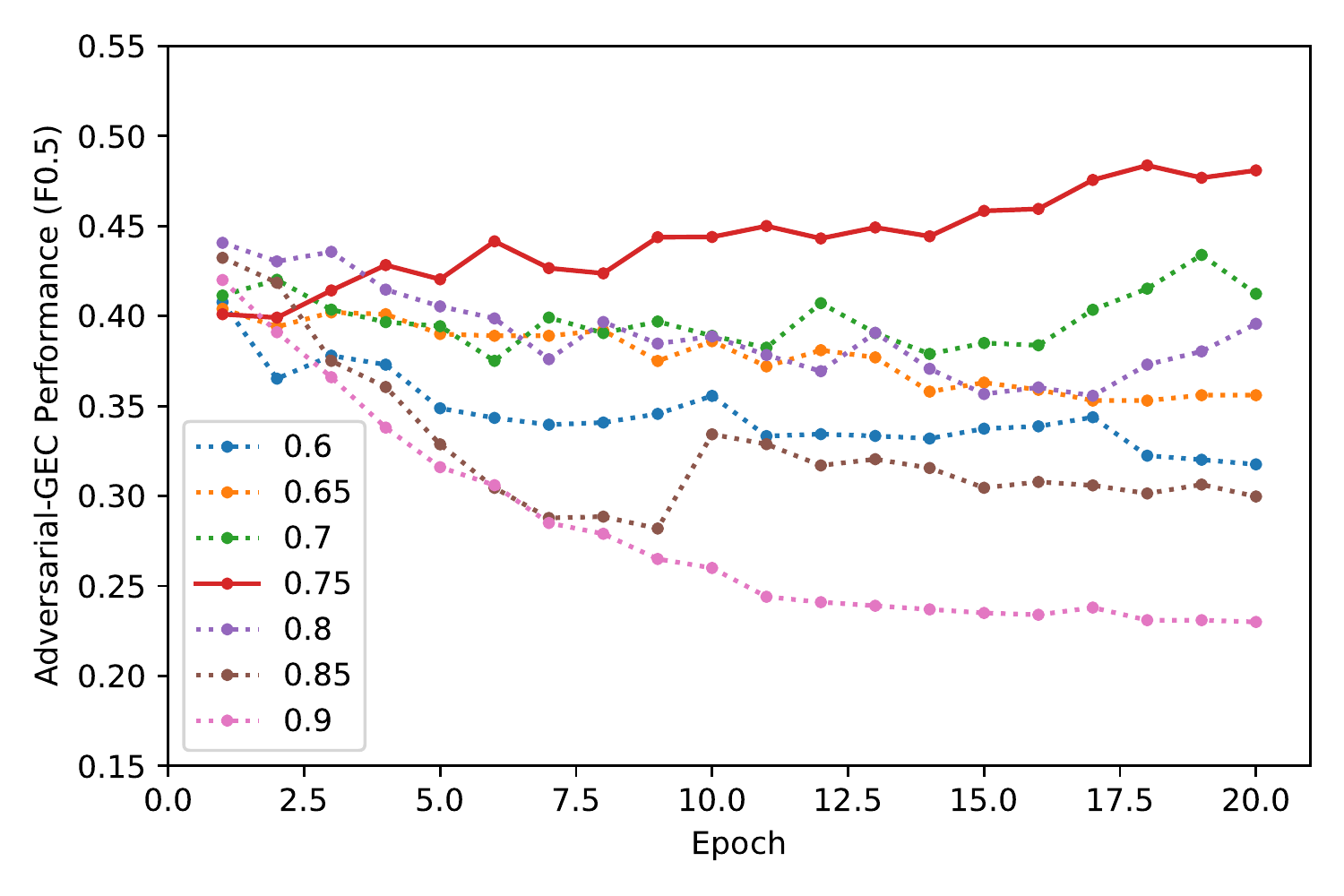}
\caption{$F_{0.5}$ scores on the dev set using pre-trained Transformer, and CNN discriminators with varying initial accuracy $\varepsilon$.}
\label{dis-pretrain}
\end{figure}

Since GAN training is a \textit{min-max} loss optimization with alternating updates to the generator and the discriminator, it is hard to reach a global optimum, which is a saddle point. To successfully reach the saddle point, balancing the generator and the discriminator co-training is essential. But the discriminator usually converges faster than the generator, so it is hard to achieve that balance. Failure to do so often leads to problems like mode collapse or inability to learn altogether. While the generator is pre-trained to reach the best development-set performance, we control the discriminator pre-training to balance the adversarial training. Hence, we evaluate the impact of the pre-trained discriminator’s accuracy $\varepsilon$ as a tunable hyperparameter. We pre-train seven RNN discriminators to reach accuracy in the range $[0.6,0.9]$. With these discriminators, we train corresponding Adversarial-GEC models (using a Transformer generator, $\lambda$ = 0.4) and evaluate their performance on the development set at regular intervals. Fig. \ref{dis-pretrain} shows that the initial accuracy of the discriminator significantly impacts the final performance and needs to be set carefully. If it is either too high (0.85 and 0.9) or too low (0.6 and 0.65), the model performs poorly. This points to the need for a balanced relationship between the generator and the discriminator. If the discriminator is too strong, the generator is extremely penalized for its erroneous predictions, and the performance progressively gets worse. On the other hand, if the discriminator is too weak, it is unable to give the most appropriate guidance to the generator. Empirically, we pre-train the discriminator until its accuracy reaches the 0.7-0.75 range.

\subsection{Combining MLE and Adversarial Objectives}
\label{combining-mle-adversarial}

\begin{figure}
\centering
\includegraphics[width=\linewidth]{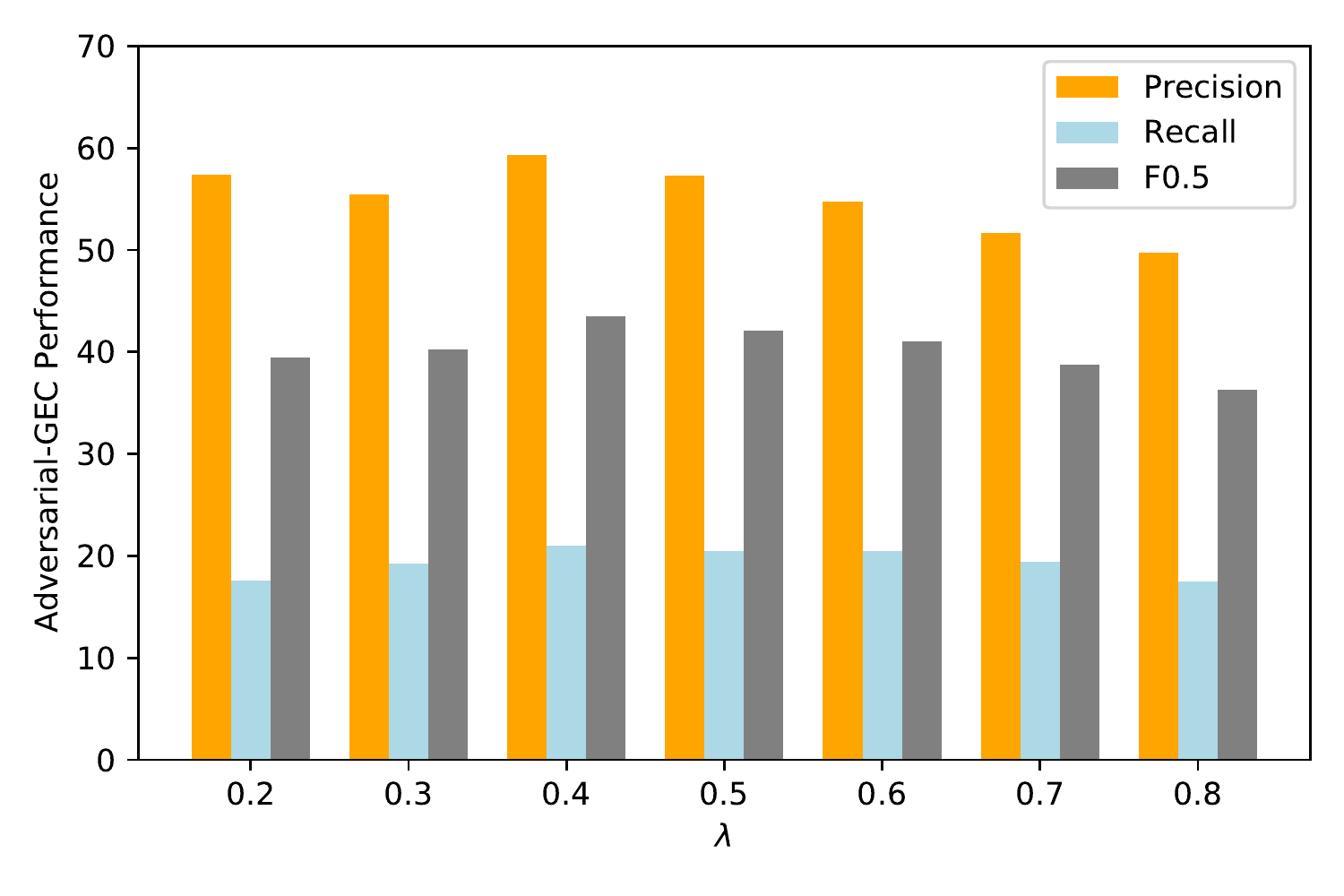}
\vspace{-0.5cm}
\caption{Adversarial-GEC performance on the dev set (Transformer + CNN), varying parameter $\lambda$ to alternate between MLE and Policy Gradient training.}
\label{dis-adv-comb}
\end{figure}

As noted in Section \ref{training-strategies}, a key factor in successfully training $G_\theta$ is the combination of adversarial and MLE objectives where we define the hyperparameter $\lambda$ to control the trade-off between MLE and adversarial training. That is, for any mini-batch, determined by a probability $\lambda$, $G_\theta$ is optimized by the MLE objective or adversarial objective to improve the stability in model training. We experiment with the range $[0.2, 0.8]$ for $\lambda$. The results in Fig. \ref{dis-adv-comb} show that combining the MLE objective with the adversarial objective is helpful to stabilize the training and improve the model performance, as we expected. This confirms prior findings that MLE acts as a regularizer to guarantee smooth model updates, alleviating the negative effects brought by high gradient estimation variance of the one-step Monte-Carlo sample in REINFORCE. However, further increasing $\lambda$ does not bring more gain. The best trade-off between MLE and adversarial objective in our experiment is $\lambda$~=~0.4, which is the value we use in our experiments. 




\subsection{Experiments with Language Models}
\label{appendix-lm}
In the \textsc{SS} setting, we also experimented with a locally-normalized language model as a discriminator. The intuition here was that using a language model with token-level locally normalized probabilities could offer a more direct training signal to the generator. If a generated sentence does not match the distribution of ground-truth data, it will have high perplexity when evaluated by a language model that was trained on ground-truth data. Not only can it provide an overall evaluation score for the whole sentence, but can also assign a probability to each token, thus providing more information on which word is to blame if the overall perplexity is very high. However, in spite of all the training strategies described in Section \ref{training-strategies}, training a language model was highly unstable, due to the use of a single sample to approximate the expected gradient, leading to high variance in gradient estimates. In future works, we aim to explore this idea using better generator models and better, larger-scale language models such as BERT \cite{devlin2018bert} and GPT-3 \cite{brown2020language}. 

\section{Related Work}

While the choice of a sentence-pair discriminator is close to \citet{yang-etal-2018-improving} and \citet{pmlr-v95-wu18a}, our work differs from \citet{yang-etal-2018-improving} in that their learning objective is a combination of the discriminator reward ($D$) and a smoothed sentence-level BLEU \cite{papineni-etal-2002-bleu} as the static reward ($Q$). The use of a sentence-pair discriminator is related to our work, we do not combine rewards from $D$ and $Q$. Incorporating $Q$ in the objective stems from the motivation to directly optimize for the evaluation metric, we choose to not force the evaluation metric-based reward into the objective, since most GEC metrics are reference-based, and have shown to be limiting for the task \cite{choshen-abend-2018-inherent, chollampatt-ng-2018-reassessment}. 
Similarly, among existing works for GEC, our work is the closest to \citet{sakaguchi-etal-2017-grammatical}, but they also directly maximize GLEU in training their GEC system, using a REINFORCE-based approach similar to ours. We instead let the model learn the latent nuances of the objective directly from the data, and provide the appropriate reward to the generator, preserving the learning objective as in \citet{yu2017seqgan}, albeit with a different discriminator framework. Our work is closest to \citet{pmlr-v95-wu18a}, who built an RNNSearch-based Generator \cite{bahdanau14nmt} and a CNN-based sentence-pair discriminator for NMT.

\section{Conclusion}
We propose a task-appropriate training objective for GEC, using an adversarial training framework consisting of a generator and a discriminator, based on the Adversarial-NMT framework of \citet{pmlr-v95-wu18a}. 
The generator is modeled as a Seq2Seq model, and the discriminator is modeled as a deep sentence-pair matching model, which provides rewards to the generator input-output. 
The framework supervises the generator to reflect the mapping within (source, target) sentence, and an efficient policy gradient algorithm to tackle the optimization difficulty brought by the discrete nature of generation. 
Experiments on standard GEC test datasets demonstrate the effectiveness of our framework for the task. 
Additionally, we provide insights into how the discriminator setup, pre-training and integration into the framework can be optimized for stable training and better performance.
We show that the proposed framework consistently achieves better results in a self-contained single model setting, without relying on any external resources.
In the future, we plan to improve the task-specific framework and training techniques based on recent state-of-the-art methods \cite{grundkiewicz-etal-2019-neural, choe-etal-2019-neural}, and improve issues with sparse rewards by exploring better credit assignment techniques.

\section*{Acknowledgments}
We would like to thank our friends and colleagues: Vivek Kulkarni, Artem Chernodub, Kostiantyn Omelianchuk, Oleksandr Skurzhanskyi, Oleksiy Syvokon, and Chad Mills, for their insightful feedback, and the anonymous reviewers for their helpful comments. 





\bibliographystyle{acl_natbib}
\bibliography{emnlp2020}

\end{document}